\pdfoutput=1

\documentclass[11pt]{article}

\usepackage[final]{coling}

\usepackage{times}
\usepackage{latexsym}

\usepackage[T1]{fontenc}

\usepackage[utf8]{inputenc}

\usepackage{microtype}

\usepackage{inconsolata}

\usepackage{graphicx}
\usepackage{multirow}

\usepackage{float}
\usepackage{wrapfig}
\usepackage{booktabs}
\usepackage{amsmath}
\usepackage{amssymb}

\usepackage{algorithm}
\usepackage{algorithmic}

\title{Learning to Reason via Self-Iterative Process Feedback for Small Language Models}

\author{Kaiyuan Chen, Jin Wang\thanks{Corresponding author} and Xuejie Zhang \\
        School of Information Science and Engineering\\
        Yunnan University\\
        Kunming, China \\
  \texttt{chenkaiyuan@stu.ynu.edu.cn, \{wangjin, xjzhang\}@ynu.edu.cn}}

\begin{document}
\maketitle
\begin{abstract}
Small language models (SLMs) are more efficient, cost-effective, and customizable than large language models (LLMs), though they often underperform in specific areas like reasoning.
Past methods for enhancing SLMs' reasoning, such as supervised fine-tuning and distillation, often depend on costly external signals, resulting in SLMs being overly confident with limited supervision signals, thus limiting their abilities.
Therefore, this study enables SLMs to learn to reason from self-iterative feedback.
By combining odds ratio preference optimization (ORPO), we fine-tune and align SLMs using positive and negative signals generated by themselves.
Additionally, we introduce process supervision for rewards in preference alignment by sampling-based inference simulation and process reward models.
Compared to Supervised Fine-Tuning (SFT), our method improves the performance of Gemma-2B by 12.43 (Acc) on GSM8K and 3.95 (Pass@1) on MBPP.
Furthermore, the proposed method also demonstrated superior out-of-domain generalization capabilities on MMLU\_Math and HumanEval.
\end{abstract}

\section{Introduction}
\label{sec:section1}
Reasoning is a popular area of research in natural language processing.
Recent studies \cite{wei2022chain, DBLP:conf/nips/KojimaGRMI22, DBLP:conf/iclr/0002WSLCNCZ23} show that closed-source LLMs with Chain-of-Thought (CoT) demonstrated excellent performance in various reasoning tasks,
even without additional fine-tuning on specialized supervised datasets \cite{cobbe2021training}.
However, due to the limited parameters, open-source SLMs ($\le$7B) haven't fully demonstrated this capability before despite being more cost-effective than their larger counterparts \cite{DBLP:conf/acl/MagisterMAMS23, DBLP:conf/acl/HoSY23}.
Therefore, enhancing the incremental reasoning abilities of SLMs has recently become a significant research focus.

\begin{figure*}[!t]
  \centering
  \includegraphics[width=0.75\linewidth]{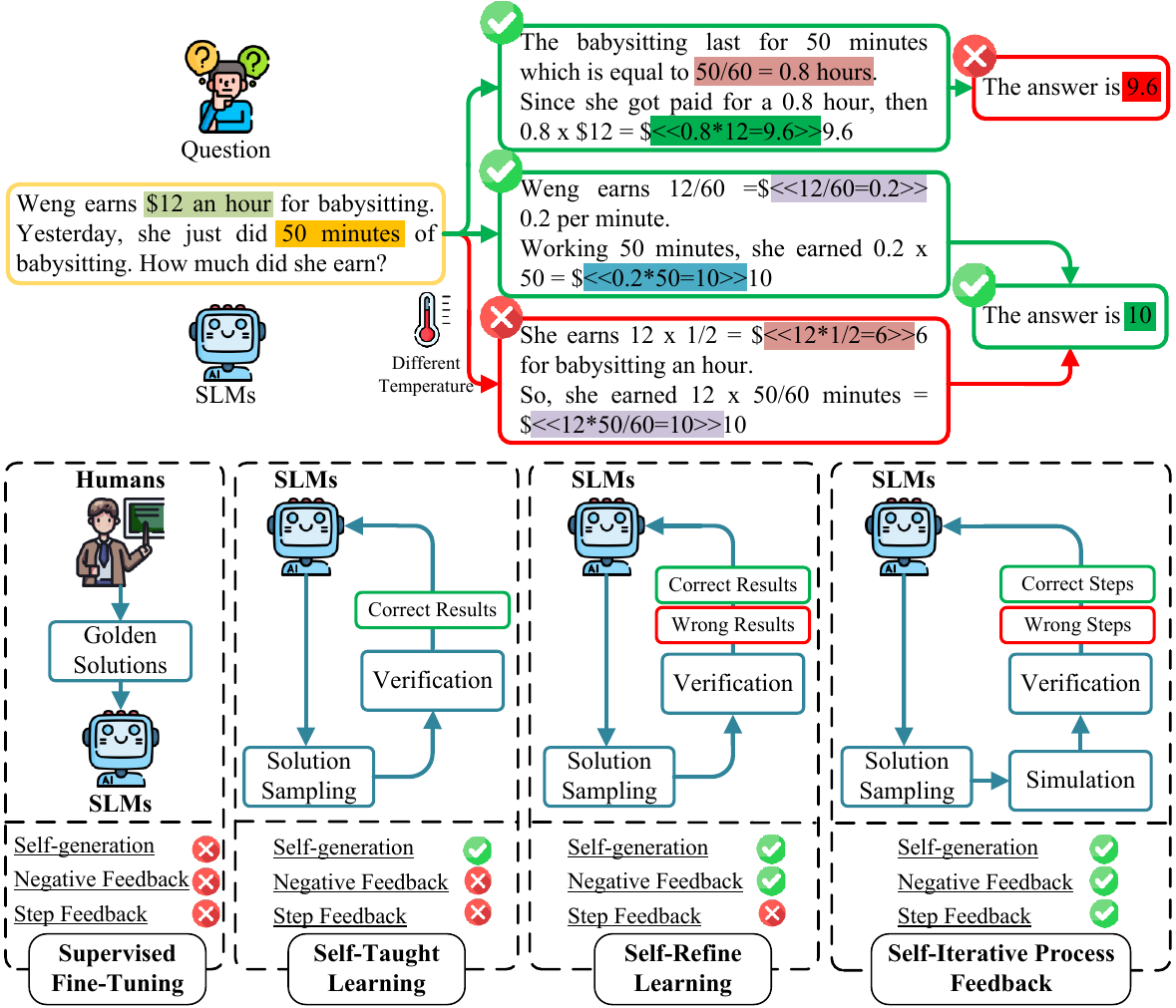}
  \caption{The conceptual diagram of the proposed self-iterative process feedback method against several previous methods (below).
  Compared to prior approaches, SIPF emphasizes on the correctness of reasoning steps.
  This means that SIPF can distinguish between correct reasoning with incorrect results and incorrect reasoning with correct results (above).}
  \label{fig:figure1}
\end{figure*}

Previous works \cite{cobbe2021training, DBLP:conf/nips/HendrycksBKABTS21,DBLP:conf/acl/MagisterMAMS23,DBLP:conf/acl/HoSY23,DBLP:conf/acl/HsiehLYNFRKLP23, chen2024mathematical} have shown that fine-tuning SLMs on meticulously designed large-scale supervised datasets effectively narrows the reasoning performance gap between open-source SLMs and closed-source LLMs.
However, constructing such datasets requires extensive supervision from humans or advanced LLMs.
Additionally, due to limited parameters and datasets, SLMs tend to become overly confident with limited supervision signals, which means that SLMs are prone to lacking generalization \cite{yuan2023scaling} or being misled by incorrect reasoning paths \cite{DBLP:conf/acl/WangM0S0Z023, bentham2024chain}, particularly when reasoning by CoT, as shown in Figure \ref{fig:figure1} (above).

Self-taught methods like STaR \cite{zelikman2022star} and RFT \cite{yuan2023scaling} make it possible to enhance language models' reasoning abilities without external annotation signals, where the model learns from self-generated reasoning paths that lead to correct final answers.
By incorporating the Direct Preference Optimization (DPO) \cite{DBLP:conf/nips/RafailovSMMEF23}, self-refine methods also aim to enhance LLMs' reasoning abilities by preference alignment on self-generated positive and negative samples, where each sample is labeled as positive or negative either by advanced LLMs (like GPT-4) \cite{lee2024reinforcement} or automatically based on whether the final answer is correct \cite{pang2024iterative}.
However, these methods depend only on result-based binary rewards, which means they overlook a significant amount of detailed step-by-step feedback, as shown in Figure \ref{fig:figure1} (below).
Please refer to Appendix \ref{sec:appendixD} for more details of related works.

To combine the advantages of self-iterative learning and fine-grained process feedback, we propose fine-tuning SLMs on self-generated positive and negative samples through self-iterative process feedback (SIPF).
A process reward model (PRM) \cite{DBLP:conf/iclr/LightmanKBEBLLS24} produces the process feedback signals without human annotations.

The proposed self-iterative method involves these steps:
First, using fine-tuned SLMs for sampling reasoning paths (CoT);
Second, applying sampling-based inference simulation to label the correctness of steps in some examples, which are then used for training the verifier (or PRM);
Next, scoring all sampled reasoning paths with the verifier and constructing preference datasets;
Finally, performing odds ratio preference optimization (ORPO) \cite{DBLP:journals/corr/abs-2403-07691} to align SLMs on preference datasets.

Unlike recent process feedback-based reasoning optimization methods \cite{jiao-etal-2024-learning,DBLP:conf/acl/WangLSXDLCWS24}, our approach focuses on gradually improving the SLMs' reasoning abilities through self-iteration.
Overall, the main contributions of this work can be summarized as follows:

\begin{itemize}
\item [1)] 
We propose a self-iterative process feedback optimization method to gradually improve reasoning in open-source SLMs without additional human-annotated signals.
It also does not need process supervision signals from humans when combined with sampling-based inference simulation.
\item [2)]
We propose improving self-refine methods by incorporating fine-grained process feedback, with process feedback rewards assigned by a PRM.
Comprehensive and rigorous analysis results indicate that incorporating process feedback leads to more robust improvements in self-refine methods.
\item [3)]
We conduct experiments on various types and scales of SLMs, including TinyLlama-v1.1, Phi-1.5, and Gemma-2B, to validate the universality of the proposed method.
Extensive experiments show that our method outperforms supervised fine-tuning, self-taught, and self-refine methods on multi-step reasoning tasks like GSM8K and MBPP.
It also achieves superior out-of-domain generalization on MMLU\_Math and HumanEval.
Additional experimental analyses also strongly support the reliability of the proposed approach.
\end{itemize}

\begin{figure*}[!t]
  \centering
  \includegraphics[width=0.9\linewidth]{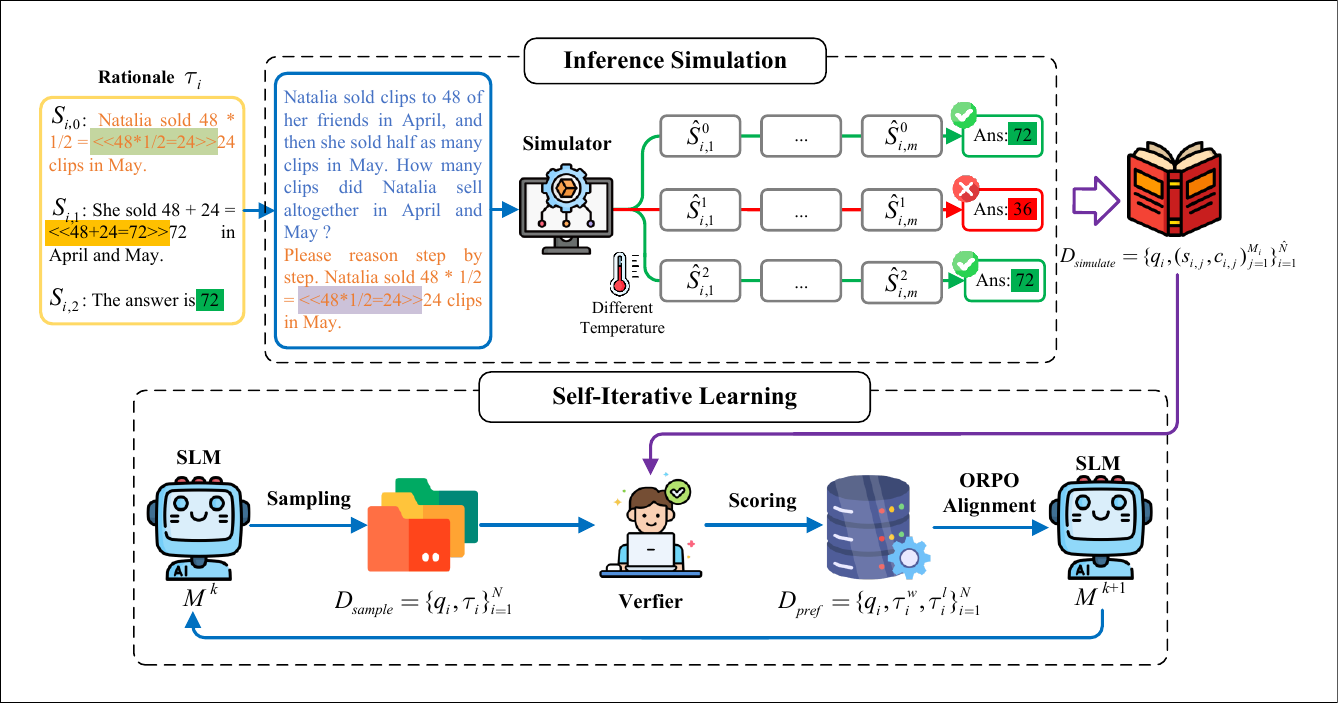}
  \caption{Overall framework of the proposed learning to reason from self-iterative process feedback.
  A single iteration of the online learning process includes sampling, collecting, inference simulation, fine-tuning verifier, scoring to construct the preference dataset, and RL alignment by ORPO.}
  \label{fig:figure2}
\end{figure*}

\section{Self-Iterative Process Feedback}
Instead of observing the correctness of the final answer, the proposed method enhances the reasoning ability of the SLMs $M$ with parameters $\theta$ by observing the correctness of each reasoning step.

We introduce a simulator $S$ and a verifier $V$, both well-trained LLMs on the specific task to achieve this goal.
Specifically, a continuous value from 0 and to 1 represents the correctness of intermediate reasoning steps.

The simulator $S$ uses the SLM's intermediate steps as input to perform multiple reasoning simulations, determining the correctness of each step based on the final results of multiple simulations.
The verifier $V$ rates the correctness scores of each step generated by the SLM and generator.
Figure \ref{fig:figure2} shows the detailed architecture of the proposed SIPF method. 

\subsection{Reasoning Paths Sampling}
The fine-tuned $M$ with parameters $\theta$ can sample a set of different reasoning paths, i.e. $\{ {\tau_0},{\tau_1},...,{\tau_n}\}$, by high temperature $T$ for a given question $q$.
The process can be formally described as follows:
\begin{equation}
R = \{ {\tau _i}|{\tau _i} \sim M({q_i},T,\theta )\} _{i = 1}^N
\end{equation}
where $R$ represents possible reasoning paths generated by $M$ given the question $q_i$ and temperature $T$.
The initial model $M_0$ is obtained by fine-tuning the original model on the original dataset $D_0$, and the $k$-th iteration model $M_k$ is obtained by aligning it on the k-th iteration dataset $D_k$ via ORPO.
Additionally, $R=\{{\tau_0},{\tau_1},...,{\tau_n}\}$ will be deduplicated and diversified based on reasoning paths (for math) or edit-distance (for code).

\subsection{Process Reward Estimation}
Compared to outcome-based rewards, which solely rely on the correctness of the final answer $a$ of reasoning.
Process supervision rewards assess the correctness of each intermediate step $s_{j}$ in the reasoning path ${\tau} = ({s_{0}},{s_{1}},...,{s_{m}},{a})$, where $j\in[0,m]$ and a is the final result.

Inspired by Monte Carlo Tree Search (MCTS), sampling-based inference simulation provides a feasible approach for determining the correctness of steps of the reasoning path without cumbersome human annotations \cite{DBLP:conf/iclr/LightmanKBEBLLS24}. 
The basic idea is that an intermediate step will likely be correct if it frequently reaches the correct result.

As shown in Figure \ref{fig:figure2}, the simulator performs multiple samplings from the reasoning step ${s_{i,j}}$ in ${\tau_{i}}$,
obtaining $K$ sampled paths: $\{ \hat \tau _{i,j}^t\} _{t = 1}^K = \{ ({s_{i,0}},...,{s_{i,j}},\hat s_{i,j + 1}^t,\hat s_{i,j + 2}^t,...\hat s_{i,m}^t,\hat a_i^t)\} _{t = 1}^K$.
The correctness ${c_{i,j}}$ of step ${s_{i,j}}$ can be defined as follows:
\begin{equation}
c_{i,j}=
\begin{cases}
        1,& \text{if } \sum \mathbb{I}(\hat a_i^{t} = a_i^*) > \delta \\
        0,& \text{otherwise}
\end{cases}
\end{equation}
where $\delta$ is threshold that denotes whether a step is considered correct only if the number of simulated correct results exceeds a specific count,
$\mathbb{I}$ represents the $0/1$ indicator function
and ${a_i^*}$ is the gold answer.

\subsection{Process Reward Model}
\label{sec:section2.3}
The verifier can be used to assign fine-grained step-level scores to reasoning paths. Given a dataset annotated through inference simulation:
\begin{align}
{D_{simulate}} = & \{ {q_i},{\tau_i},{c_i}\} _{i = 1}^{\hat N}
\notag
\\= & \{ {q_i},({s_{i,j}},{c_{i,j}})_{j = 1}^{{M_i}}\} _{i = 1}^{\hat N}
\end{align}
where ${M_i}$ represents the total number of steps in ${\tau_i}$ and $\hat{N}$ represents the number of samples less than $N$. The corresponding loss function can be formalized as follows:
\begin{align}
  {{\cal L}_{PRM}} = & - \sum\limits_{i = 0}^N \sum\limits_{j = 0}^{{M_i}} 
      {c_{i,j}}\log {{\hat c}_{i,j}} \notag\\
      & + (1 - {c_{i,j}})\log (1 - {{\hat c}_{i,j}})
\end{align}
where ${\hat c_{ij}}$ is the probability assigned by the verifier that step ${s_{i,j}}$ is correct, which is ranges from $[0,1]$.
Therefore, the reward $r$ of an reasoning path ${\tau} = ({s_{0}},{s_{1}},...,{s_{m}},{a})$ of the i-th question ${q_i}$ can be estimated by the fine-tuned verifier as follows:
\begin{align}
r{\rm{(}}\tau {\rm{)}} = \frac{{\sum\nolimits_{j = 0}^m {{{\hat c}_{i,m}}} }}{m}
\end{align}
where $m$ is the total number of reasoning steps.
Since the performance of verification is related to language modeling capabilities,
the verifier is usually an open-source model with strong reasoning performance.

\subsection{RL Alignment via ORPO}
Based on the PRM, we can assign different reward $r$ to various reasoning paths $\tau$, thereby constructing the preference dataset as follows:
\begin{equation}
{D_{pref}} = \{ {q_i},{\tau _i}^w,{\tau _i}^l\left| {r({\tau _i}^w) - r({\tau _i}^l) \ge \eta } \right.\}
\end{equation}
where ${\tau^w}$ is the chosen path while ${\tau^l}$ is the reject one and $\eta$ is the confidence margin.
Benefiting from the efficiency of ORPO, supervised fine-tuning and preference alignment can be incorporate into a single process.
It allows the model to learn more diverse reasoning from the chosen reasoning paths and avoid erroneous patterns through preference alignment.
The learning objectives corresponding to this process are denoted as:

\begin{algorithm}
  \caption{Iterative Training Procedure}
  \label{alg:algorithm1}
  \begin{algorithmic}[1]
  \STATE \textbf{Initialize:} pretrained SLM $M$; fine-tuned verifier $V$; dataset $D_0 = \left\{ {{q_i},{\tau_i},{a_i}} \right\}_{i = 1}^N$; 
  \STATE $M_{0} = \text{SFT}(M, D_{0})$
  \FOR{$k = 1$ to $N$}
      \STATE $D_{\text{sample}} = \text{Sample}(M_{k-1}, D_{k-1})$
      \STATE $D_k = D_{k-1} \cup D_{\text{sample}}$
      \STATE $D_{\text{pref}} = \text{Score}(V, D_k)$
      \STATE $M_k = \text{ORPO}(M, D_{\text{pref}})$
  \ENDFOR
  \end{algorithmic}
\end{algorithm}

\begin{align}
\mathcal{L}_{ORPO} &= -\mathbb{E}_{(q,\tau^{w},\tau^{l})}\Bigg[
  \mathcal{L}_{SFT}(q,\tau^{w})  \notag\\
  & \quad + \beta \left( \log \sigma \left( \frac{\mathrm{odds}(\tau^{w}\mid q)}{\mathrm{odds}(\tau^{l}\mid q)} \right) \right) 
\Bigg]
\end{align}
\begin{equation}
{\rm{odds(}}\tau \left| q \right.{\rm{)}} = \frac{{{P}(\tau \left| q \right.)}}{{1 - {P}(\tau \left| q \right.)}}
\end{equation}
where ${{\cal L}_{SFT}}(q,{\tau^w})$ is is the negative log-likelihood loss on ${\tau^w}$, $\sigma$ is log sigmoid function,
$\beta$ is weight parameters, and ${\rm{odds}}(\tau \left| q \right.)$ represents the ratio between the probability of obtaining reasoning path $\tau$ given question $q$ and the probability of not generating it.  

\subsection{Self-Iterative Process Feedback}
Through self-iteration, SLMs and datasets can be updated iteratively, overcoming the limitations of finite datasets and reducing bias with continuous online feedback.
This process involves a series of iterative models ${M_0},{M_1},...,{M_k}$ and their corresponding datasets ${D_0},{D_1},...,{D_k}$.
Algorithm \ref{alg:algorithm1} shows the details of self-iterative process feedback.

\begin{table*}[!t]
	\centering
   {\fontsize{10}{12}\selectfont
    \begin{tabular}{ccccc}
		\toprule
		Model &\multicolumn{2}{c}{TinyLlama-v1.1} &\multicolumn{2}{c}{Gemma-2B} \\
		\midrule
		Method\textbackslash Task   &GSM8K   &MMLU\_Math (OOD)   &GSM8K  &MMLU\_Math (OOD) \\
    \midrule
    CoT (8-shot)   &1.14   &5.08  &14.40  &13.28 \\
    SFT            &13.75  &6.21  &31.54  &24.01 \\
    \midrule
    RFT            &14.40  &6.21  &34.80  &24.58 \\
    pRFT           &14.56  &7.34  &34.57  &28.25 \\
    SRF            &7.43   &2.26  &27.98  &23.73 \\
    pSRF           &9.63   &3.95  &28.35  &23.73 \\
    RPO            &14.25  &6.21  &37.91  &27.12 \\
    STaR           &14.63  &5.65  &34.72  &24.58 \\
    \midrule
    SIPF-Iter1     &16.22  &7.34  &38.15  &25.98 \\
    SIPF-Iter2     &17.44  &7.91  &42.00  &\textbf{29.10} \\
    SIPF-Iter3     &\textbf{18.57}  &\textbf{9.60}  &\textbf{43.97}  &28.25 \\
    \bottomrule
	\end{tabular}
  }
  \caption{Comparison of different methods on the mathematical reasoning tasks GSM8K (in-domain) and MMLU\_Math (out-of-domain).}
  \label{tab:table1}
\end{table*}

\section{Experiments}

\subsection{Datasets}
The experiments involve two challenging types of multi-step reasoning tasks: mathematical reasoning (GSM8K \cite{cobbe2021training}, MMLU\_Math \cite{DBLP:conf/iclr/HendrycksBBZMSS21}) and code generation (MBPP \cite{austin2021program}, HumanEval \cite{chen2021evaluating}).

GSM8K involves solving math word problems in grade school. MBPP tasks involve basic programming knowledge, requiring multi-step reasoning and passing unit tests. Both datasets offer publicly available training and testing sets.

To assess the models' out-of-domain generalization, we will train them on GSM8K and MBPP, and then evaluate them on MMLU\_Math and HumanEval respectively.
MMLU\_Math comprises college, high school, and elementary school mathematical problems from MMLU.
Please refer to Appendix \ref{sec:appendixA.1} for more details of datasets.

\subsection{Experiment Setup}
The experiments consider three types and scales of pre-trained SLMs, including TinyLlama-v1.1 \cite{zhang2024tinyllama}, Phi-1.5 \cite{li2023textbooks}, and Gemma-2B \cite{team2024gemma}.
All models in the experiments are fine-tuned by QLoRA \cite{DBLP:conf/nips/DettmersPHZ23} and employ Unsloth\footnote{\url{https://github.com/unslothai/unsloth}} for inference acceleration.

The deepseek-math-7b-instruct and deepseek-math-7b-rl \cite{shao2024deepseekmath} are employed as the simulator and reward model for mathematical reasoning, respectively, and deepseek-coder-6.7b-instruct \cite{guo2024deepseek} as both the simulator and reward model for code generation.

In each iteration, we collect sufficient positive and negative samples through temperature sampling and ensure sample diversity by deduplicating based on reasoning paths (for math) or edit-distance (for code).
Inference simulation is performed by sampling eight times for all tasks, with temperatures set to 1 for math tasks and 0.7 for code tasks.
Please refer to Appendix \ref{sec:appendixA.2} for more setup details.

\begin{table*}[!t]
	\centering
    {\fontsize{10}{12}\selectfont
    \begin{tabular}{ccccc}
		\toprule
		Model &\multicolumn{2}{c}{Phi-1.5} &\multicolumn{2}{c}{Gemma-2B} \\
		\midrule
		Method\textbackslash Task   &MBPP   &HumanEval (OOD)   &MBPP  &HumanEval (OOD) \\
    \midrule
    CoT (0-shot)   &34.87  &33.54  &28.29  &20.12 \\
    SFT            &35.20  &32.32  &29.75  &17.68 \\
    \midrule
    RFT            &34.60  &26.22  &27.49  &14.02 \\
    pRFT           &36.60  &28.05  &28.56  &15.24 \\
    SRF            &22.38  &17.07  &16.47  &12.80 \\
    pSRF           &25.92  &19.51  &19.17  &11.59 \\
    RPO            &28.52  &33.54  &31.53  &20.73 \\
    STaR           &\textbf{38.44}  &29.88  &28.56  &15.85 \\
    \midrule
    SIPF-Iter1     &36.87  &\textbf{34.76}  &31.83  &20.12 \\
    SIPF-Iter2     &38.21  &29.88  &32.16  &\textbf{21.95} \\
    SIPF-Iter3     &37.14  &31.10  &\textbf{33.70}  &20.73 \\
    \bottomrule
	\end{tabular}
  }
  \caption{Comparison of different methods on the code generation tasks MBPP (in-domain) and HumanEval (out-of-domain).}
  \label{tab:table2}
\end{table*}

\subsection{Baseline}
In addition to CoT and supervised fine-tuning, we consider the following existing methods as strong baselines: \\
\textbf{Self-Taught Methods:} STaR \cite{zelikman2022star} and RFT \cite{yuan2023scaling} are Self-Taught methods based on outcome feedback, where the models only learn from the reasoning paths that lead to correct results.
RFT boosts diversity in self-generated reasoning with high-temperature sampling, while STaR uses greedy decoding and iteratively improves performance. \\
\textbf{Self-Refine Methods:} Similar to the approach used by \citet{yuan2024self,lee2024reinforcement}, we consider using SFT and DPO to align the model with self-sampled positive and negative examples.
However, labels of examples are not assigned by LLMs but are instead automatically assigned based on the correctness of the final results.
In subsequent experiments, this baseline will be referred to as SRF.
Additionally, RPO \cite{pang2024iterative} optimizes reasoning by aligning self-generated positive and negative outcome feedback.
We compare RPO's performance after one iteration in the experiments.

For a strict comparison, we use PRMs described in Section \ref{sec:section2.3} to introduce process feedback for RFT and SRF.
The improved methods are referred to as pRFT and pSRF, respectively.
Please refer to Appendix \ref{sec:appendixA.3} for more baselines details.

\subsection{Evaluation}
We evaluate the model's accuracy for mathematical reasoning tasks, while for code generation, Pass@1 is used as the evaluation metric.
By default, all results are generated using greedy decoding by the model.
For more details on the evaluation, please refer to Appendix \ref{sec:appendixA.1}.

\section{Results and Analysis}

\subsection{Comparison of In-Domain Performance}
Tables \ref{tab:table1} and \ref{tab:table2} indicate that SIPF demonstrates superior in-domain performance on multi-step reasoning tasks: mathematics (GSM8K) and code generation (MBPP).
Specifically, SIPF shows significant improvements over SFT and CoT.
This improvement is observed across different models.

With just one iteration, SIPF's in-domain performance surpasses self-taught and self-refine methods and their process feedback-based versions (pRFT and pSRF).
Additionally, with the help of self-iteration, SIPF achieves continuous improvements in most cases, as shown in Tables \ref{tab:table1} and \ref{tab:table2}.

\subsection{Comparison of Out-of-Domain Performance}
As mentioned in Section \ref{sec:section1}, most methods based on external signals are constrained by limited resources and fail to exhibit good out-of-domain performance.
As indicated in Table \ref{tab:table2}, SFT shows degraded performance in the code generation task HumanEval (out-of-domain) compared to CoT.

In contrast, in most cases, SIPF demonstrates strong out-of-domain performance, surpassing existing methods like STaR, RFT, and SRF.
SIPF also further improve its performance on MMLU\_Math through self-iteration, as shown in Table \ref{tab:table1}.
However, for the code generation task HumanEval, the improvements from self-iteration are limited, as shown in Table \ref{tab:table2}.

\subsection{Analysis of Performance Degradation in DPO-based Self-Refine Methods}
\begin{figure*}[t]
  \includegraphics[width=0.48\linewidth]{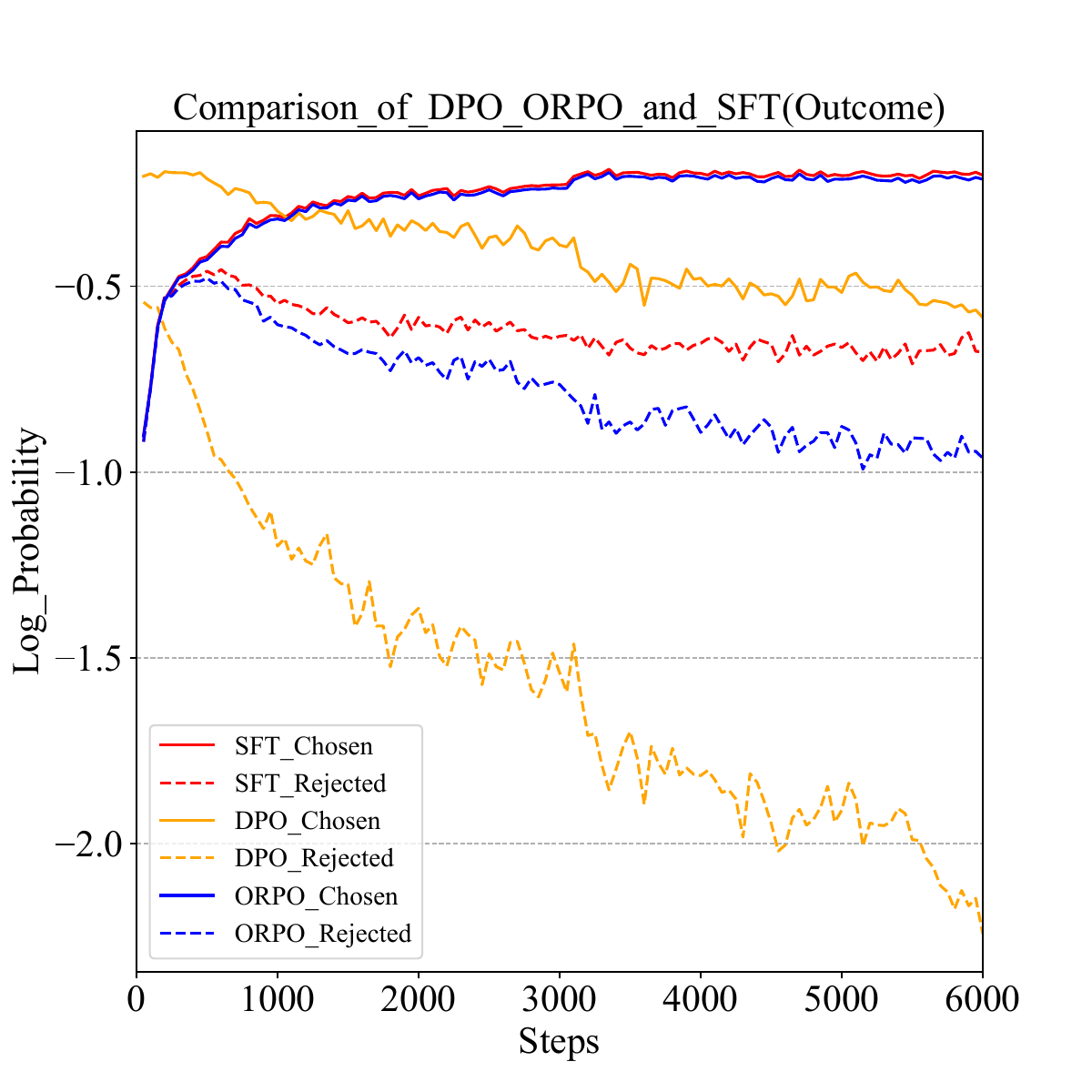} \hfill
  \includegraphics[width=0.48\linewidth]{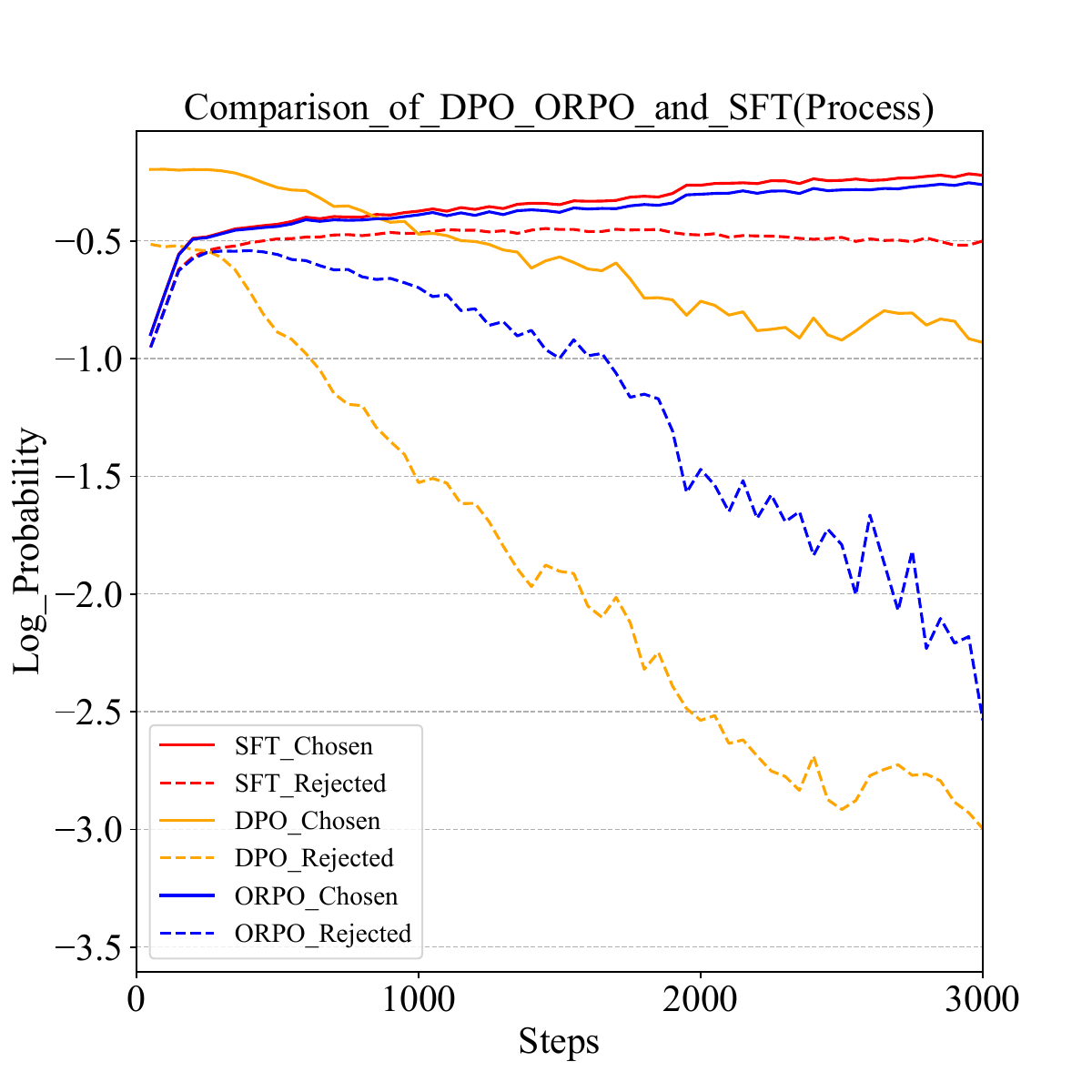}
  \caption {Comparison of the training processes of SFT-based self-taught, DPO-based SRF, and ORPO-based SIPF on GSM8K using Gemma-2B.
  The DPO-based SRF consistently reduced the probability of generating ${\tau^w}$.
  SIPF is more effective than self-taught methods at reducing the probability of generating ${\tau^l}$.
  Additionally, different types of feedback (outcome or process) have different effects on training.}
  \label{fig:figure3}
\end{figure*}

Tables \ref{tab:table1} and \ref{tab:table2} indicate that SRF shows degraded performance compared to SFT in all cases, regardless of whether it is based on outcome or process feedback.
As shown in Figure \ref{fig:figure3}, we analyze the chosen and rejected probability across different models and feedback types to further explore the underlying reasons.

Figure \ref{fig:figure3} shows that DPO reduces the probability of generating ${\tau^w}$ during training, which directly causes performance degradation in SFT models after aligning feedback signals.
These results are consistent with findings from previous works \cite{DBLP:journals/corr/abs-2403-07691,pang2024iterative}.
In fact, the loss form of DPO causes the model to focus excessively on what is a poor response (as shown in Figure \ref{fig:figure3}) while overlooking what makes a good response. It has also been confirmed in recent work \cite{feng2024towards}.
In contrast, ORPO-based SIPF improves the probability of generating ${\tau^w}$ due to the supervised fine-tuning on ${\tau^w}$.

\subsection{Analysis of Aligning Positive and Negative Examples}
Self-Taught methods focus solely on iterative supervised fine-tuning on self-generated positive examples, while SIPF further incorporates aligning on positive and negative examples via ORPO.

Figure \ref{fig:figure3} shows that SFT increases the generation probability of ${\tau^w}$ while it does not effectively reduce the probability of ${\tau_l}$, which is consistent with the observations of previous works.
In comparison, SIPF further lowers the likelihood of generating rejected examples, indicating that it more effectively reduces erroneous reasoning in SLMs and improves reasoning performance.
Additionally, different types of feedback affect SFT and ORPO differently.
The rejected probability curves of SFT and ORPO differ more significantly from process feedback to outcome feedback.
Under outcome feedback, both SFT and OPRO reduce the probability of generating rejected samples, leading to a similar performance in handling rejected samples.

\begin{figure}[!t]
  \includegraphics[width=\columnwidth]{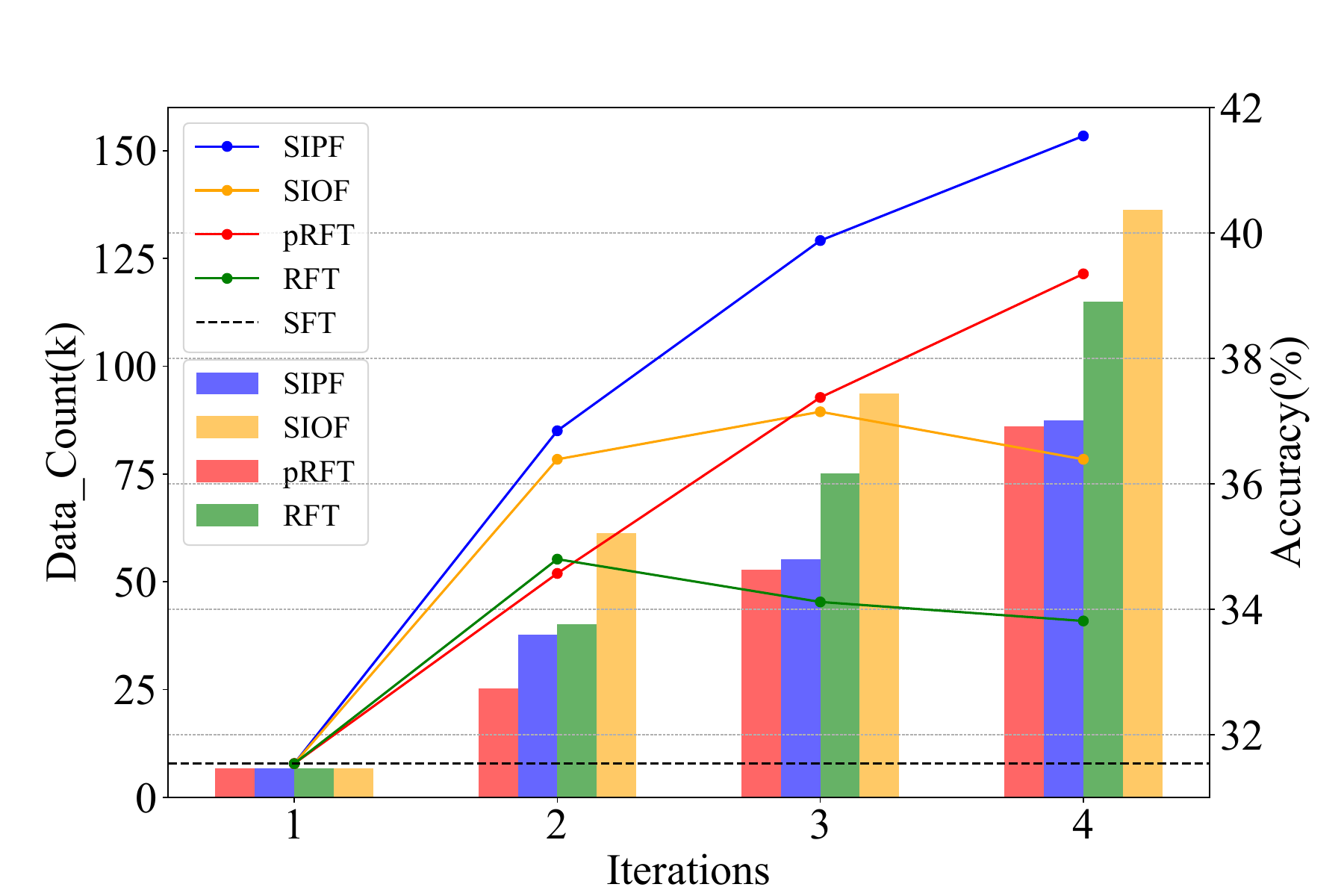}
  \caption{Comparison of accuracy and data count for various self-iteration methods across different iterations on GSM8K.}
  \label{fig:figure4}
\end{figure}

\begin{figure}[!t]
  \includegraphics[width=\columnwidth]{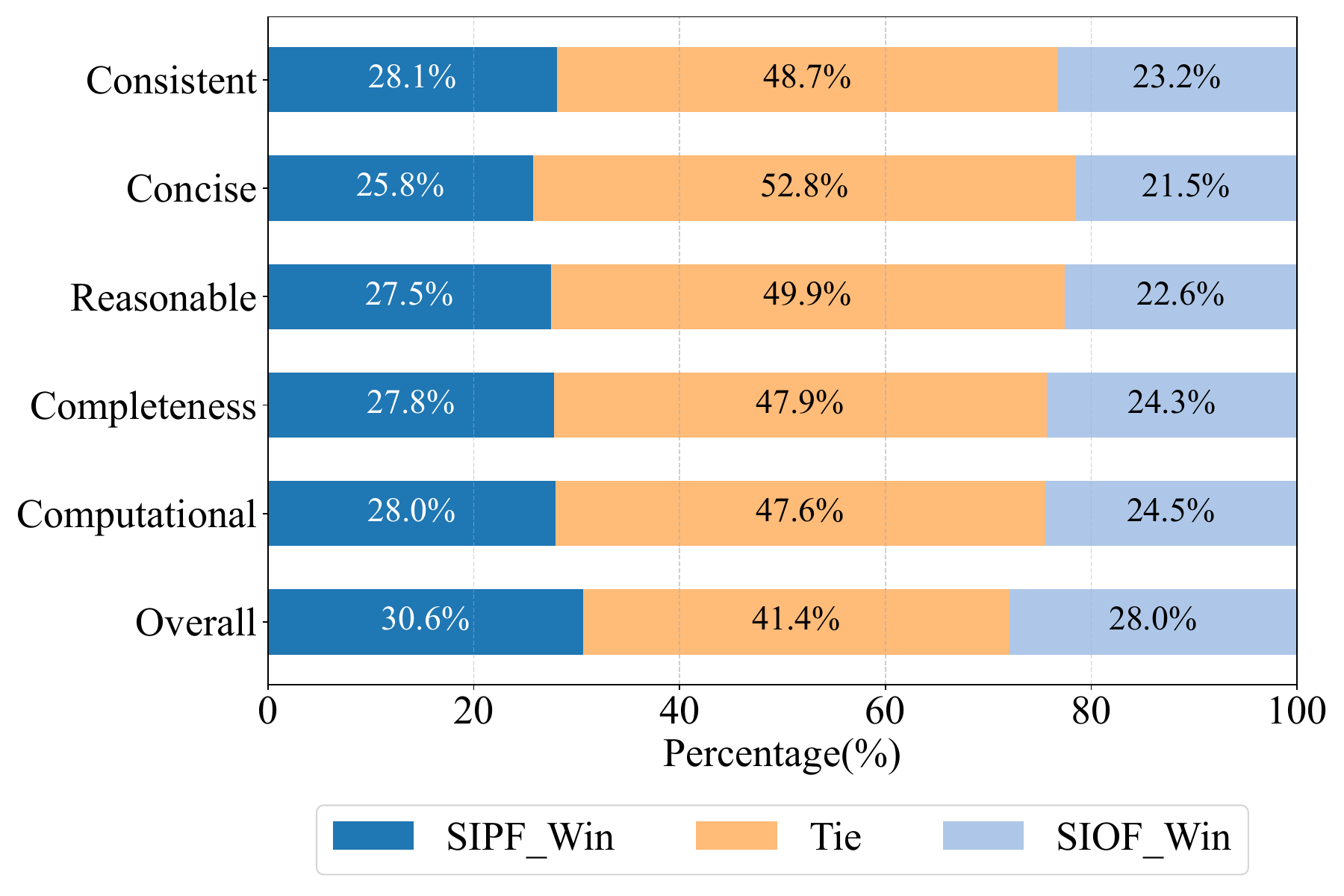}
  \caption{Automatic evaluation of the reliability of rationales generated by different methods (process-based or outcome-based) using GPT-4.}
  \label{fig:figure6}
\end{figure}

\subsection{Comparison of Different Self-Iterative Learning Methods}
\label{sec:section4.5}
We compare the accuracy and data count changes during continuous iteration for four self-iterative methods based on process or outcome feedback.
Specifically, RFT can also be implemented as a self-iterative optimization method, similar to \cite{DBLP:journals/tmlr/SinghCAAPGLH0XP24}.
SIOF represents iterative learning using only outcome feedback, with other settings consistent with SIPF.

Figure \ref{fig:figure4} shows that all methods achieve higher accuracy than the SFT baseline through self-iteration learning.
SIPF with process feedback signals consistently outperforms other methods in every iteration.

It is worth noting that the differing impacts of process feedback and outcome feedback on reasoning performance become evident through multiple iterations.
RFT and SIOF, based on outcome feedback, show degraded performance after reaching their capacity limits.
In contrast, process feedback methods (SIPF and RFT) continue to improve, likely due to higher-quality examples.
This is also reflected in the observed data count changes across iterations, where more training examples do not always lead to better reasoning performance.

Additionally, aligning positive and negative samples benefits reasoning performance in iterative learning. Figure 5 shows that SIPF and pRFT outperform SIOF and RFT, respectively.

\begin{table}[!t]
	\centering
    {\fontsize{10}{12}\selectfont
    \begin{tabular}{lcc}
		\toprule
		Method\textbackslash Type    &ORM   &PRM \\
    \midrule
    TinyLlama-v1.1 &59.64  &\textbf{67.00} \\
    Llama-2-7B      &70.09  &\textbf{78.38} \\
    deepseek-math-instruct-7B  &75.60  &\textbf{85.24} \\
    deepseek-math-rl-7B        &77.38  &\textbf{87.58} \\
    \bottomrule
	\end{tabular}
  }
  \caption{Accuracy analysis of different reward models trained with process and result feedback on manually annotated GSM8K evaluation dataset.}
  \label{tab:table3}
\end{table}

\subsection{Effect of Process Feedback on Reasoning}
To explore whether process feedback can effectively reduce reasoning errors and improve the reasoning process, we follow \citet{zheng2023judging} and use GPT-4 to automatically evaluate the rationales generated under different methods. 

The GPT-4 is required to analyze the reliability of the reasoning process from six different perspectives: \textit{Computational}, \textit{Completeness}, \textit{Reasonable}, \textit{Concise}, \textit{Consistent}, and \textit{Overall Performance}. 

Specifically, we prompt GPT-4 to compare and evaluate the rationales generated by SIPF and SIOF (as described in Section \ref{sec:section4.5}) to determine which method produces more reliable paths.
To eliminate the influence of the final answer, we removed examples where the result is incorrect under both methods.
The results in Figure \ref{fig:figure6} show that process-based SIPF outperforms outcome-based SIOF methods across multiple dimensions of rationales evaluation.
For more details on the evaluation, please refer to Appendix \ref{sec:appendixB}.

\subsection{Analysis of Performance across Different Reward Models}
As described in Section \ref{sec:section2.3}, the fine-tuned reward models can be used to assign reward scores to each step in the rationales, thus constructing a reference dataset.
Therefore, the accuracy of the reward model is crucial for enhancing SIPF's performance.
However, recent works have shown a lack of analysis concerning the reward models.

To evaluate the performance of the reward model, we manually labeled a dataset specifically for assessing step correctness.
Following \citet{DBLP:conf/iclr/LightmanKBEBLLS24}, we train PRMs with step-annotated rationales and ORMs with result-annotated rationales.
The fine-tuned reward models then assign correctness scores to each step in the evaluation dataset to assess accuracy.
Please refer to Appendix \ref{sec:appendixC} for more details on manual annotations.

Table \ref{tab:table3} shows that, compared to ORM, PRM can more accurately evaluate the correctness of reasoning steps, consistent with previous work \cite{DBLP:conf/iclr/LightmanKBEBLLS24}.
Additionally, the performance of the reward model is closely related to its language modeling capabilities \cite{cobbe2021training}.
The stronger the model's reasoning ability, the better it performs in evaluations.
Specifically, models like deepseek-math-rl-7B and deepseek-math-instruct-7B, which are aligned using large-scale, high-quality mathematical datasets, significantly outperform TinyLlama-v1.1 and Llama-2-7B \cite{touvron2023llama}, especially in the case of PRM.

\begin{table}[!t]
	\centering
    {\fontsize{10}{12}\selectfont
    \begin{tabular}{lcc}
		\toprule
		Method\textbackslash Model    &Gemma-2B   &TinyLlama-v1.1 \\
    \midrule
    SIPF    &38.15   &16.22 \\
    \midrule
    \textit{w ITR}    &\textbf{43.97}  &\textbf{18.57} \\
    \textit{w/o PF}   &37.83  &15.24 \\
    \textit{w/o OR}   &34.64  &14.63 \\
    \textit{w/o PFOR} &34.57  &14.40 \\
    \bottomrule
	\end{tabular}
  }
  \caption{Ablation study on GSM8K (Accuarcy). ITR denotes self-iterative learning (3 rounds),
  \textit{PF} represents process feedback, \textit{OR} refers to the relative ratio loss in the ORPO objective, and \textit{PFOR} represents both process feedback and relative ratio loss.}
  \label{tab:table4}
\end{table}

\section{Ablation Analysis}
\begin{figure}[!t]
  \includegraphics[width=\columnwidth]{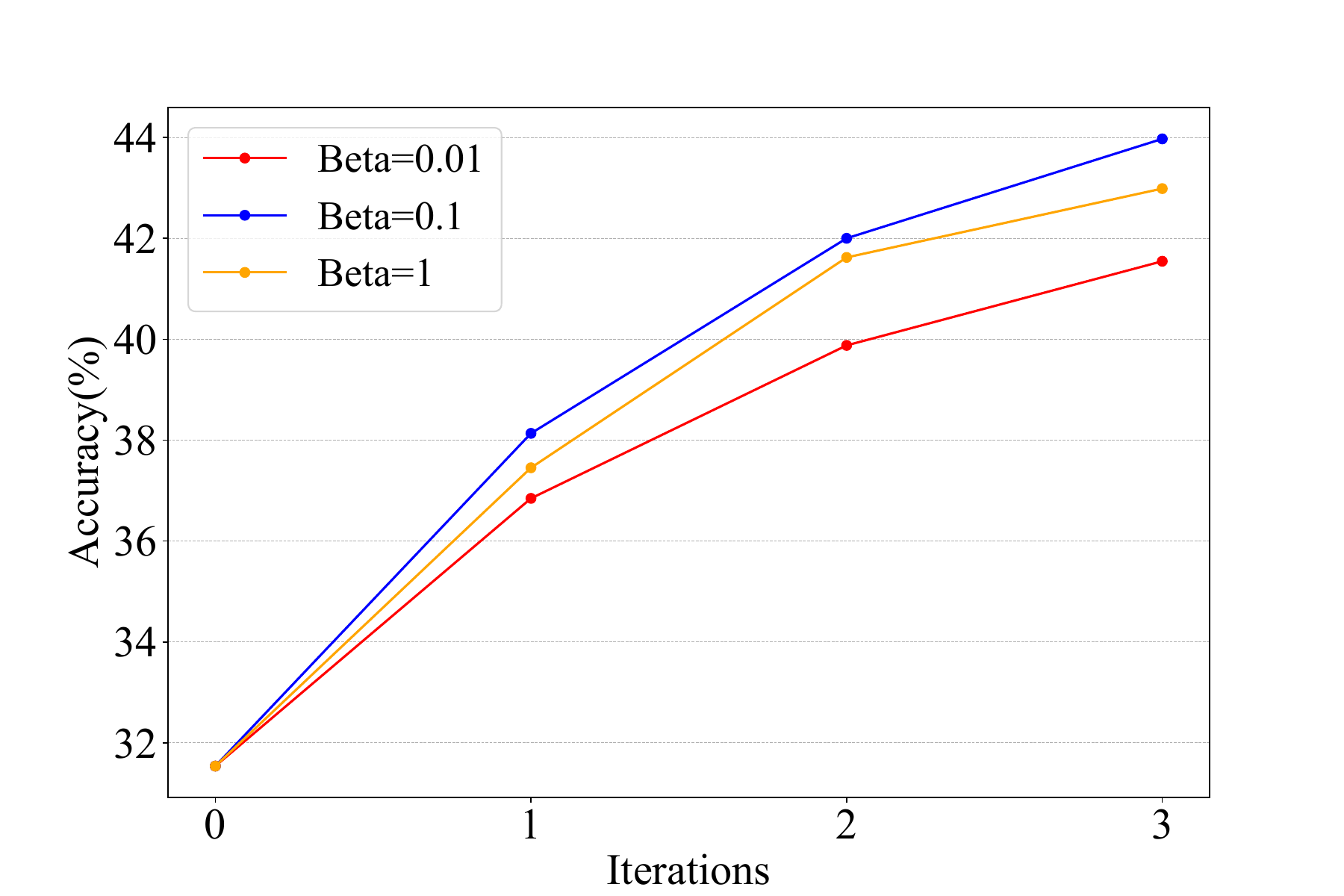}
  \caption{Comparison of performance with different weights of relative ratio loss across various iterations on GSM8K using Gemma-2B.}  
  \label{fig:figure5}
\end{figure}

We conduct an ablation analysis on different components of the proposed method to validate their effectiveness.
Table \ref{tab:table4} presents the ablation experiment results of Gemma-2B and TinyLlama-v1.1 on the GSM8K task, with SIPF using only one iteration as the reference. 
It shows that performance degrades across different models when various components are removed, highlighting the necessity of each part. We also explore the performance differences with varying weight of relative ratio loss, as shown in Figure \ref{fig:figure5}.
The results indicate that an OR weight of around 0.1 is suitable for SIPF.

\section{Conclusion}
This work introduces a self-iterative process feedback method aimed at improving the reasoning abilities of SLMs, 
where process feedback is generated through reasoning simulation and verification, without relying on manual annotation.
Extensive experimental results across multiple multi-step reasoning tasks demonstrate the effectiveness and generalizability of the proposed method.

\section*{Limitations}
The proposed method has only been experimented on pre-trained models with sizes $\le$ 2B, and its feasibility has not been validated on larger open-source models ($\ge$ 7B).
Current experimental results only demonstrate the method's effectiveness on common multi-step reasoning tasks (e.g., mathematics and code) and do not confirm its applicability to a broader range of reasoning tasks.
Additionally, the resource overhead of the self-iterative approach itself is significant and cannot be overlooked.

\section*{Acknowledgements}
This work was supported by the National Natural Science Foundation of China (NSFC) under Grant Nos. 61966038 and 62266051, and the Postgraduate Practice and Innovation Foundation of Yunnan University under Grant No. ZC-242410094.
We would like to thank the anonymous reviewers for their constructive comments.

\bibliography{custom}


\appendix
\section{Experimental Details}
\label{sec:appendixA}
\subsection{Dataset}
\label{sec:appendixA.1}
\textbf{GSM8K.} GSM8K \cite{cobbe2021training} is a challenging multi-step math reasoning task, where each sample consists of grade school math word problems annotated with step-by-step solutions.
It contains 7,473 samples as the training set, while the test set consists of 1,319 samples. \\
\textbf{MMLU\_Math.} MMLU \cite{DBLP:conf/iclr/HendrycksBBZMSS21} is a benchmark task designed to evaluate a model's ability to solve problems by applying world knowledge across various domains, including mathematics, law, and computer science.
The MMLU\_Math used in our experiments is a subset of MMLU. Specifically, it includes college, high school, and elementary school math problems from MMLU.
Since all questions in the original task are multiple-choice, we only retain those where the correct answer is a standard number.
After filtering, only 354 samples are used as the test set. \\
\textbf{MBPP.} MBPP \cite{austin2021program} consists of 1,000 entry-level Python programming problems.
Each problem includes a task prompt, a code solution, and 3 test cases.
We use a training set containing 374 samples and a test set with 500 samples in the experiments.
Notably, because different combinations of test cases can be included in the prompts, the actual number of training samples used for fine-tuning is 2,244.
For testing, to prevent label leakage, we selected 2 out of 3 test cases as prompts for each problem, resulting in 6 different input formats per question.
Under these conditions, we evaluated the performance on MBPP using Pass@k metrics. \\
\textbf{HumanEval.} HumanEval \cite{chen2021evaluating} contains 164 entry-level programming problems to assess understanding, arithmetic, and reasoning.
We use this dataset to evaluate the model's out-of-domain generalization capabilities on code generation.
Additionally, to maintain consistency with the format of MBPP, we transform the input format of Humaneval problems.
Specifically, the input for a problem is changed from \textit{Function\_Name}+\textit{Prompt}+\textit{Test\_Case} to \textit{Prompt}+\textit{Test\_Case}+\textit{Function\_Name}.

\begin{table}[!t]
	\centering
    {\fontsize{10}{12}\selectfont
    \begin{tabular}{lc}
		\toprule
		Hyperparameters    &Setting  \\
    \midrule
    learning\_rate    &5e-5, 8e-5, 1e-4  \\
    optimizer        &AdamW  \\
    warmup\_ratio     &0.1   \\
    max\_grad\_norm    &0.3  \\
    OR\_weight        &0.01, 0.1, 1  \\
    bf16             &True  \\
    lora\_alpha       &128  \\
    lora\_dropout     &0.05  \\
    lora\_r           &256   \\
    \bottomrule
	\end{tabular}
  }
  \caption{Unified parameter settings for self-iteration alignment across different tasks.}
  \label{tab:table5}
\end{table}

\begin{figure*}[t]
  \includegraphics[width=0.48\linewidth]{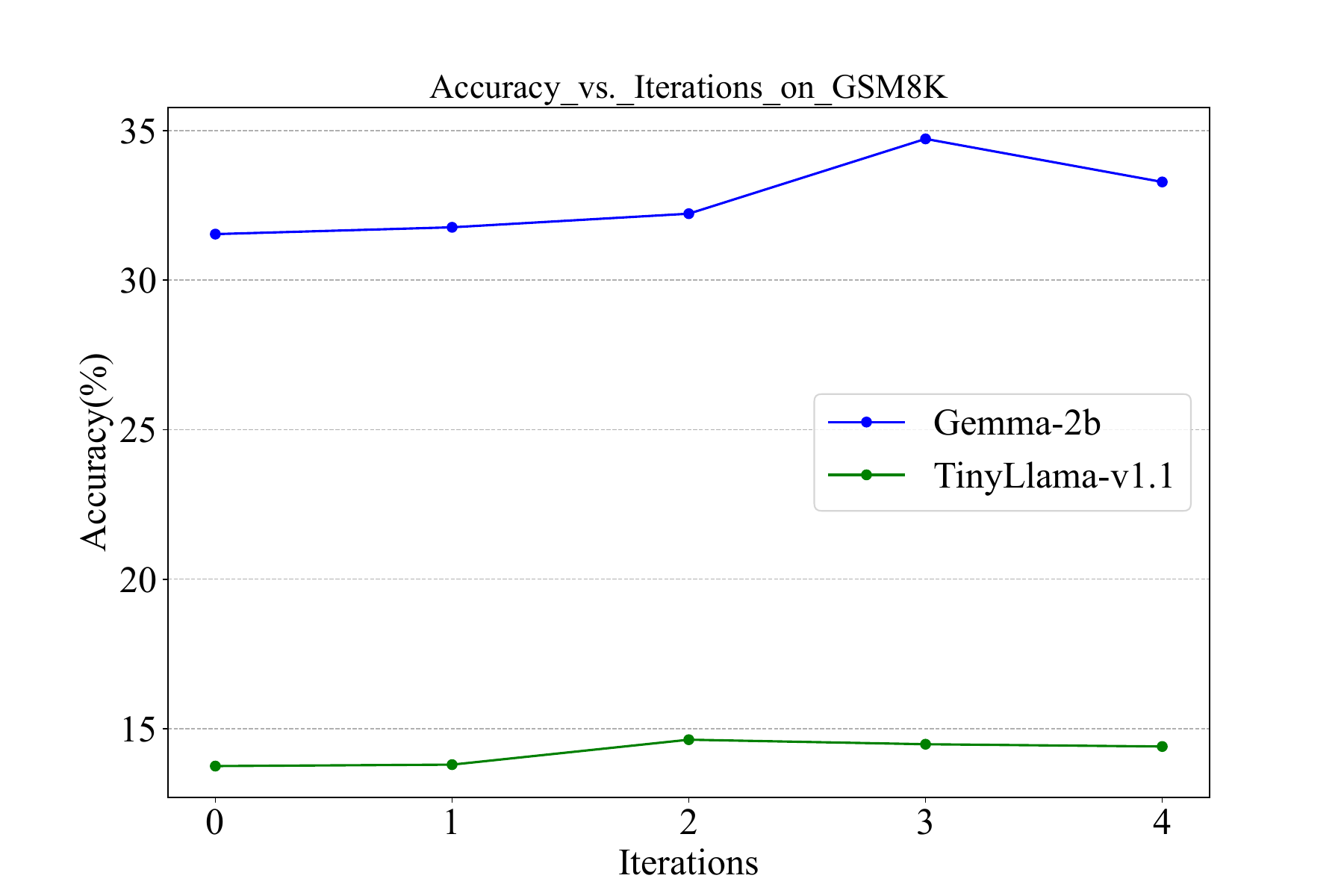} \hfill
  \includegraphics[width=0.48\linewidth]{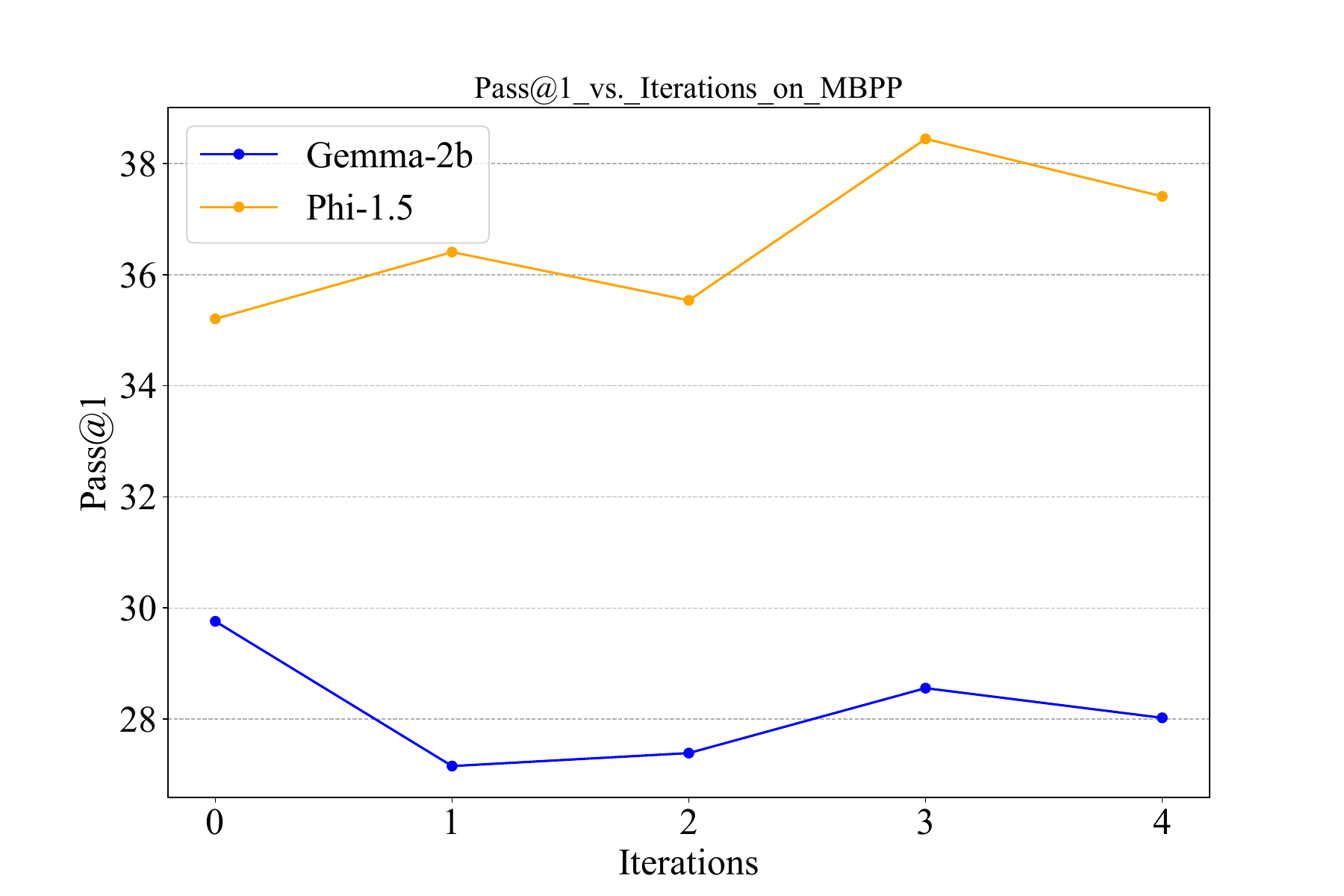}
  \caption {The performance of STaR at different iteration rounds on GSM8K (Acc) and MBPP (Pass@1).
  Iteration-0 represents supervised fine-tuning on the original dataset.}
  \label{fig:figure7}
\end{figure*}

\subsection{Experiment Setup}
\label{sec:appendixA.2}
\textbf{Training.} For efficiency considerations, we use Unsloth in our experiments and apply QLoRA for parameter-efficient fine-tuning.
All methods are based on algorithms and models implemented in the transformer \cite{wolf2020transformers} and trl\footnote{\url{https://github.com/huggingface/trl}} libraries.
Table \ref{tab:table5} shows the hyperparameter settings used for different tasks and models.
When training Gemma-2B, we set the batch size to 20, while the batch size is 48 for TinyLlama-v1.1 and Phi-1.5.
Due to the complexity of the self-iteration learning, we select the best performance under a predefined set of learning rates and relative ratio loss weights from Table 5 
and leave the search for optimal parameters for future work.
We set the batch size to 40 for training the reward model and used a low learning rate 5e-5. \\
\textbf{Sampling.} In each iteration, temperature sampling ($T = 1$) wil be used to collect enough positive and negative samples (reasoning paths),
labeling them based on the correctness of the final answer for math problems or test passes for program problems.
Specifically, we specify the required number of positive and negative samples per question.
Based on subsequent rounds, the number will be set to half or a quarter of the number of positive and negative samples from previous rounds.
Since not all math problems yield the required number of samples through sampling, we set a time threshold for sampling to improve efficiency.
For code generation problems, samples are labeled post-sampling, so no time threshold is set. 
Tables \ref{tab:table7} and \ref{tab:table6} show the count of sampled datasets obtained after filtering at different iteration rounds.

\begin{table}[!t]
	\centering
    {\fontsize{10}{12}\selectfont
    \begin{tabular}{lcc}
		\toprule
		Iteration\textbackslash Model    &Gemma-2B   &Phi-1.5 \\
    \midrule
    Iter-0    &2.24  &2.24 \\
    Iter-1    &13.12  &18.35 \\
    Iter-2    &20.06  &30.43 \\
    Iter-3    &31.27  &39.87 \\
    \bottomrule
	\end{tabular}
  }
  \caption{Training data size (k) obtained through sampling and filtering by reward values at different iteration rounds on MBPP.}
  \label{tab:table7}
\end{table}

\begin{table}[!t]
	\centering
    {\fontsize{10}{12}\selectfont
    \begin{tabular}{lcc}
		\toprule
		Iteration\textbackslash Model    &Gemma-2B   &TinyLlama-v1.1 \\
    \midrule
    Iter-0    &7.47  &7.47 \\
    Iter-1    &37.78  &36.96 \\
    Iter-2    &55.24  &50.49 \\
    Iter-3    &87.41  &68.94 \\
    \bottomrule
	\end{tabular}
  }
  \caption{Training data size (k) obtained through sampling and filtering by reward values at different iteration rounds on GSM8K.}
  \label{tab:table6}
\end{table}

\subsection{Baseline}
\label{sec:appendixA.3}
\textbf{CoT.} On GSM8K and MMLU\_Math, we used the same few-shot prompt as in the previous work \cite{wei2022chain}, while the zero-shot prompt is used for MBPP and HumanEval. \\
\textbf{SFT.} The models are fine-tuned on the existing training sets of GSM8K and MBPP. \\
\textbf{STaR.} STaR \cite{zelikman2022star} is a Self-Taught method that enhances reasoning by iteratively learning from self-generated rationales, with each rationale generated through greedy decoding.
Due to the inclusion of low-quality samples, we did not adopt the rationalization from prior work \cite{zelikman2022star}.
With a low learning rate of 5e-5, the performance of models converges after 4 iterations as shown in Figure \ref{fig:figure7},
and we selected the best performance from different iterations. \\
\textbf{RFT.} Rejection Sampling Fine-Tuning (RFT) \cite{yuan2023scaling} enhances the training dataset by self-sampling from a supervised model, with each sample filtered for uniqueness in its reasoning path.
Unlike STaR, RFT does not use self-iterative optimization.
Our experiments set the sampling temperature to 1 and increased reasoning diversity through path deduplication for all tasks. \\
\textbf{SRF.} Compared to Self-Taught, Self-Refine \cite{yuan2024self,lee2024reinforcement} focuses on correcting errors through negative feedback.
We apply DPO to align the supervised model using the first round of sampling dataset from SIPF.
Due to resource constraints, we only classify samples based on whether the final answer is correct (for math) or if it passes test cases (for code). \\
\textbf{RPO.} Iterative Reasoning Optimization (RPO) \cite{pang2024iterative} improves reasoning performance through iteration by combining negative log-likelihood loss on positive examples with a DPO learning objective on both positive and negative examples.
However, this method is still based on outcome feedback.
In the experiments, we applied RPO to the first round of sampling datasets from SIPF. \\
\textbf{pRFT and pSRF.} Original RFT and STR optimize solely based on outcome feedback.
Therefore, we use a process reward model to further filter the feedback results,
meaning only samples with high reward values are used for model optimization.

\begin{figure}[!t]
  \includegraphics[width=\columnwidth]{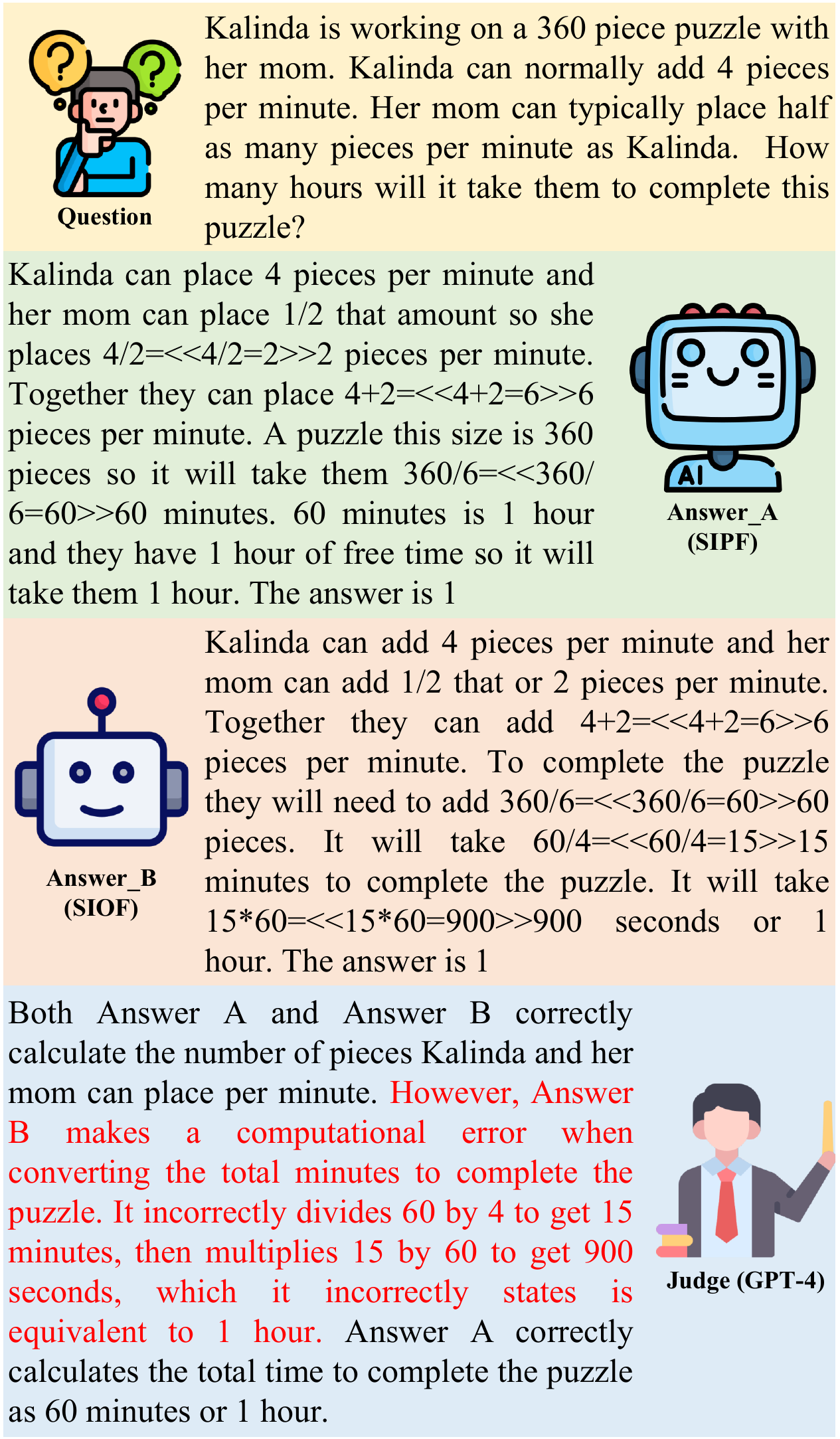}
  \caption{The case study of automatic evaluation of rationales.
  With prompting, GPT-4 identifies the reasoning error in Answer\_B, even though both Answer\_A and Answer\_B have the same correct results.}  
  \label{fig:figure9}
\end{figure}

\begin{figure*}[!t]
  \includegraphics[width=\linewidth]{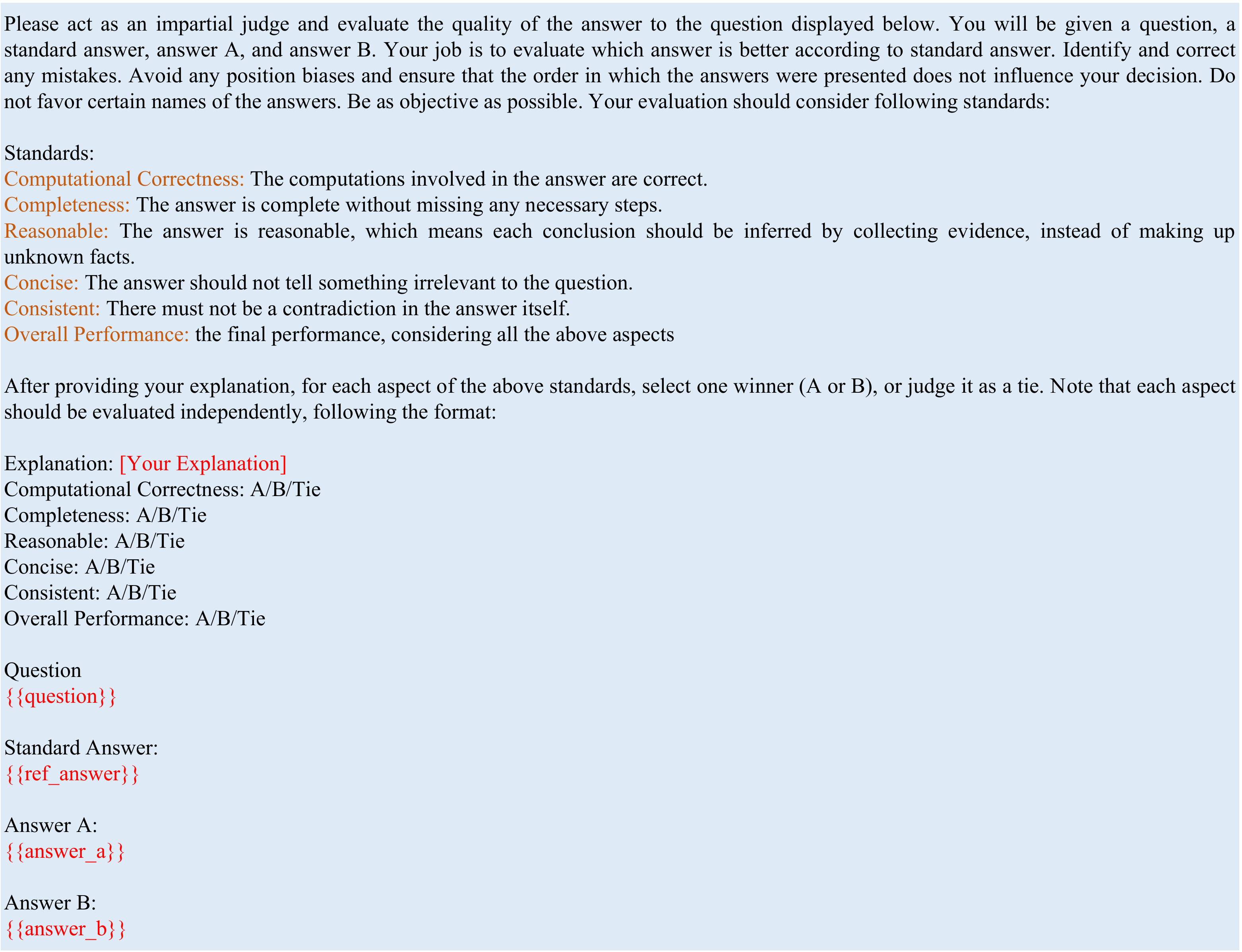}
  \caption{In the GPT-4-based automatic evaluation of rationales, the prompt templates provide definitions for six different evaluation dimensions and ask GPT-4 to determine which answer, A or B, is better or equivalent, after providing the relevant explanations.}  
  \label{fig:figure8}
\end{figure*}

\section{Automatic Evaluation of Rationales}
\label{sec:appendixB}
Due to the advanced capabilities, powerful LLMs like GPT-4 can align closely with human preferences, allowing them to serve as judges in various evaluation tasks \cite{zheng2023judging}.
To evaluate the rationales generated by different methods, we used GPT-4 as the judge. Specifically, GPT-4 was given a question from GSM8K along with two different answers, and with appropriate prompting, it determined which response was better.
Building on the definitions of reasoning errors and validity from previous work \cite{jiao-etal-2024-learning, DBLP:conf/acl/WangM0S0Z023}, we compared the rationales across six different dimensions: Computational, Completeness, Reasonable, Concise, Consistent, and Overall Performance.
The definitions of each dimension and the prompt templates used are shown in Figure \ref{fig:figure8}. Additionally, we provide a case study of judgments from GPT-4 in Figure \ref{fig:figure9}.

\section{Manual Annotation of Rationales}
\label{sec:appendixC}
\begin{figure*}[!t]
  \includegraphics[width=\linewidth]{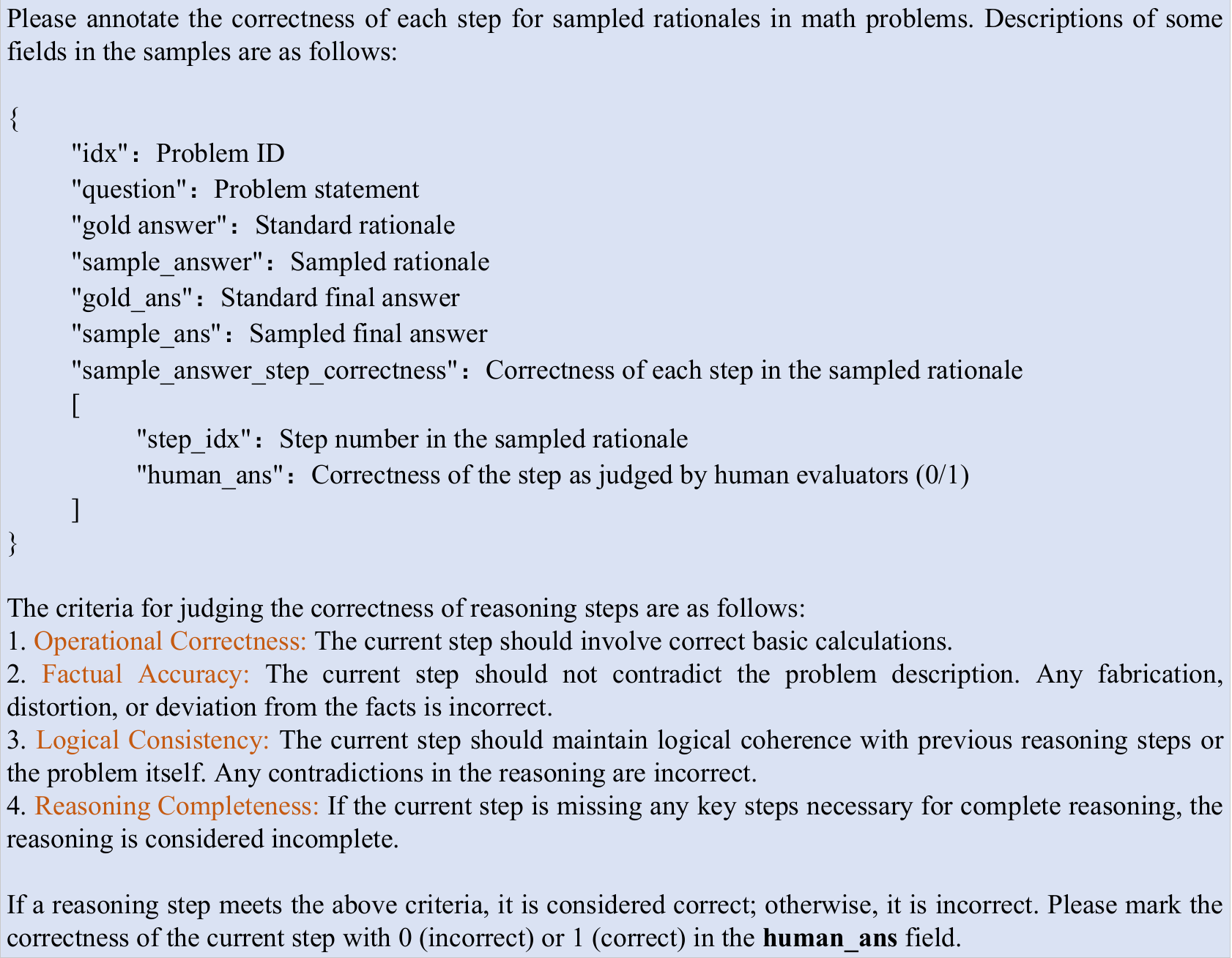}
  \caption{The guideline for annotating reasoning steps in sampled rationales from GSM8K.
  The guidelines provide four criteria for evaluating the correctness of indirect steps, including: \textit{Operational Correctness}, \textit{Factual Accuracy}, \textit{Logical Consistency}, and \textit{Reasoning Completeness}.}  
  \label{fig:figure10}
\end{figure*}

To explore the alignment between the reward model and human standards, 
we collected 1,000 rationales from GSM8K (all generated by model sampling) and distributed them to different human evaluators,
asking them to annotate the correctness of each step in the rationales.
Specifically, we provided four criteria for judging the correctness of reasoning steps, as shown in the annotation guidelines in Figure \ref{fig:figure10}.
After removing some invalid labeled samples, we retained 791 samples for evaluating the reward model.

\section{Related Work}
\label{sec:appendixD}
\textbf{Chain-of-thought Reasoning.} Previous work \cite{wei2022chain} has shown that by inserting a few Chain-of-Thought examples into the input context or by simply adding \textit{Let's think step by step} \cite{DBLP:conf/nips/KojimaGRMI22} after the question,
pre-trained LLMs can be guided to perform multi-step reasoning, thereby enhancing reasoning performance.
Recently, guiding models to perform multi-step reasoning has also been deliberately introduced into open-source language models through supervised fine-tuning \cite{cobbe2021training, chung2024scaling} or knowledge distillation \cite{DBLP:conf/acl/MagisterMAMS23, DBLP:conf/acl/HoSY23, DBLP:conf/acl/HsiehLYNFRKLP23, shen2023graphs, chen2024mathematical, shen2025knowledge}.

However, such methods often rely on expensive labeled data from humans or LLMs, which limits the scalability of the approach to some extent.
The limited data availability constrains the further enhancement of the model's capabilities \cite{chen2024self}, reducing its generalization \cite{yuan2023scaling} and potentially leading to hallucination issues \cite{rawte2023survey}. \\
\textbf{Self-Training Reasoning.} To overcome the limitations of manual annotation, STaR \cite{zelikman2022star} was the first to propose leveraging the inherent language modeling capabilities of pre-trained language models to improve reasoning.
Specifically, STaR fine-tunes the model on self-generated rationales that yield correct results, and this process is iteratively repeated several times.
Unlike STaR, RFT \cite{yuan2023scaling} believes that high-temperature sampling from pre-trained models can significantly enhance the diversity of generated rationales, and diverse reasoning approaches are crucial for improving the model's reasoning performance.
However, RFT does not emphasize improving reasoning through iterative training. Building on the previous works, ReST$^{EM}$ \cite{DBLP:journals/tmlr/SinghCAAPGLH0XP24} further describes self-training as an expectation-maximization reinforcement learning (RL) process.
These Self-Taught methods only learn from correct solutions and do not consider the impact of incorrect solutions.

Benefiting from preference alignment techniques (e.g., DPO \cite{DBLP:conf/nips/RafailovSMMEF23}),
recent works \cite{yuan2024self, chen2024self} have attempted to use self-generated positive and negative samples to improve the instruction-following capabilities of LLMs, where LLMs themselves typically assign each sample a label.
For reasoning tasks, V-STaR \cite{DBLP:journals/corr/abs-2402-06457} proposes using DPO to train a outcome-supervised verifier (or ORMs) on self-sampled positive and negative examples and combining this verifier to enhance the reasoning performance of self-iterative models.
Additionally, RPO \cite{pang2024iterative} enhances reasoning performance through iterative learning by combining negative log-likelihood loss on positive examples with a DPO learning objective that applies to both positive and negative examples.
However, these works still rely on outcome feedback. \\
\textbf{Learn Reasoning from Process Feedback.} Due to the inherent complexity of reasoning and the potential lack of fidelity in language models, 
the correctness of the result often does not align with the reasoning process \cite{DBLP:conf/icml/ZhangPMLS24, bentham2024chain}.
This means that a reasoning process that yields a correct answer may contain erroneous steps, while a reasoning process that leads to an incorrect answer can also include correct steps.
This uncertainty limits further improvements in the reasoning performance of language models.

Based on these considerations, \citet{DBLP:conf/iclr/LightmanKBEBLLS24} introduced the Process Supervision Reward Model (PRM), which is trained on reasoning paths with manually annotated step correctness and used for evaluating sampled paths.
To avoid the tedious and costly manual annotation process, recent works \cite{jiao-etal-2024-learning, DBLP:conf/acl/WangLSXDLCWS24} have used path simulation in Monte Carlo Tree Search to assign labels for the correctness of each step.
Alternatively, fine-grained feedback for the reasoning process is introduced through the self-reflective capabilities of advanced LLMs \cite{lee2024reinforcement}, such as GPT-4.
Our work is based on assigning correctness to each step through inference simulations and improving reasoning performance through a self-iterative optimization process.

\end{document}